\documentclass[a4paper, 10pt, conference]{ieeeconf}      % Use this line for a
\usepackage{FG2025}
\usepackage{amsmath}
\usepackage{url} 

\FGfinalcopy % *** Uncomment this line for the final submission

\IEEEoverridecommandlockouts                              % This command is only
\def\FGPaperID{26} % *** Enter the FG2025 Paper ID here

\title{\LARGE \bf
Measuring Anxiety Levels with Head Motion Patterns in Severe Depression Population
}

\author{\parbox{16cm}{\centering
    {\large Fouad Boutaleb$^{1,4}$, Emery Pierson$^{2}$, Nicolas Doudeau$^{4}$, Clémence Nineuil$^{4}$, Ali Amad$^{4}$, and Mohamed Daoudi$^{1,3}$}\\
    {\normalsize
    $^1$ Univ. Lille, CNRS, Centrale Lille, Institut Mines-Télécom, UMR 9189 CRIStAL, F-59000 Lille, France\\
    $^2$ LIX, École Polytechnique, IP Paris\\
    $^3$ IMT Nord Europe, Institut Mines-Télécom, Univ. Lille, Centre for Digital Systems, F-59000 Lille, France\\
    $^4$Univ. Lille, Inserm, CHU Lille, U1172 - LilNCog - Lille Neuroscience \& Cognition, F-59000 Lille, France\\
    }}
    \thanks{The French State under the France-2030 programme and the Initiative of Excellence of the University of Lille are acknowledged for the funding and support granted to the R-CDP-24-005-CALYPSO project.} % <-this % stops a space
}

\begin{document}

\thispagestyle{empty}
\pagestyle{empty}
\maketitle

%%%%%%%%%%%%%%%%%%%%%%%%%%%%%%%%%%%%%%%%%%%%%%%%%%%%%%%%%%%%%%%%%%%%%%%%%%%%%%%%
\begin{abstract} 
Depression and anxiety are prevalent mental health disorders that frequently cooccur, with anxiety significantly influencing both the manifestation and treatment of depression. An accurate assessment of anxiety levels in individuals with depression is crucial to develop effective and personalized treatment plans. This study proposes a new noninvasive method for quantifying anxiety severity by analyzing head movements -specifically speed, acceleration, and angular displacement - during video-recorded interviews with patients suffering from severe depression. Using data from a new CALYPSO Depression Dataset, we extracted head motion characteristics and applied regression analysis to predict clinically evaluated anxiety levels. Our results demonstrate a high level of precision, achieving a mean absolute error (MAE) of 0.35 in predicting the severity of psychological anxiety based on head movement patterns. This indicates that our approach can enhance the understanding of anxiety's role in depression and assist psychiatrists in refining treatment strategies for individuals.
\end{abstract}

%%%%%%%%%%%%%%%%%%%%%%%%%%%%%%%%%%%%%%%%%%%%%%%%%%%%%%%%%%%%%%%%%%%%%%%%%%%%%%%%

\section{Introduction}

Depression is a prevalent mental health disorder that affects approximately 280 million people worldwide~\cite{WHO2023}. According to the Diagnostic and Statistical Manual of Mental Disorders (DSM-5), the symptoms of Clinical Depression can manifest in many various ways, including severe depression with suicidal thoughts ~\cite{american2013diagnostic}. Objective means of detecting severe depression are crucial to enable early diagnosis and better medical treatments. Recent research has shown that multiple, non-verbal, behavioral indicators of depression can be used for this objective~\cite{depression1,depression2,depression3}.

However, depression remains a complex condition that varies significantly between individuals~\cite{allsopp2019heterogeneity}. Severely depressed patients often experience additional symptoms, such as psychomotor retardation or anxiety, which can present as physical or psychological distress. Psychiatrists frequently observe these symptoms. Evaluation of these symptoms is critical in guiding treatment decisions and supporting patient recovery~\cite{choi2020comorbid}. In other words, understanding the severity of each symptom allows clinicians to tailor treatments more effectively to each patient.

The use of behavioral indicators to provide a detailed analysis of a patient's status has not yet been fully explored. Patient examinations are heavily based on subjective evaluations, depending on the clinician's experience and the patient’s ability to communicate their symptoms~\cite{fitzgerald2017implicit}. This variability affects the reliability of diagnoses and may miss subtle but important differences in symptoms such as anxiety. Automatic detection algorithms based on non-verbal cues could help overcome these challenges by offering more consistent and objective evaluations, providing critical insights into managing severely depressed patients.

Wearable physiological devices, such as heart rate monitors or electrodermal activity sensors, have been used to detect several anxiety disorders.% ~\cite{}.
While these physiological measures provide valuable information, they are often costly and not easily accessible for widespread clinical use. Additionally, the intrusive nature of wearable devices can impact patient comfort and compliance, limiting their practicality in routine assessments.

At the same time, it has become a general fact that emotion and patients' feelings can be understood from visual media~\cite{wang2023unlocking}.
Moreover, extracting behavioral markers, such as head movements, from videos offers the advantage of being discreet and easily integrated into standard interactions, providing continuous and real-time analysis without specialized equipment. 
% To investigate this, we partnered with the University Hospital Center of X \footnote{we remove the name of the university for the double-blind review} 

To explore the possibility of obtaining personalized, adaptive treatments based on objective, non-intrusive markers, we captured the CALYPSO Depression Dataset. This dataset includes data from patients diagnosed with severe depression, along with detailed psychiatrist-assessed anxiety levels. Moreover, it contains informal interviews designed to reproduce daily life scenarios, such as meetings with a general practitioner.

%This dataset allowed us to focus on meaningful data, ensuring that the resulting model remains transparent, interpretable, and practical for real-world clinical use. 

% We propose a novel, non-invasive approach to assessing anxiety by analyzing head movements. By capturing unconscious, natural movements, this method provides objective, quantifiable data that can improve diagnostic accuracy and support more personalized treatment strategies.
Won et al.~\cite{won2016identifying} have shown that head movements are a useful indicator of anxiety. Anxiety, whether psychological or physiologic, is a common symptom of severe depression. In particular, it intensifies the effects of depression and complicates treatment~\cite{choi2020comorbid}. The presence of psychological anxiety is always reported in Hamilton interviews~\cite{hamilton1960rating, first2004structured}, making this data available for our approach. We introduce a novel method to assess whether or not, measuring anxiety levels of severely depressed patients can be done using head motion information. This method is by nature, non-invasive and analyzes head movements by separating motion and non-motion sequences during the interview. Our pipeline provides objective, measurable, and interpretable data that can improve diagnostic accuracy and support more personalized treatment strategies.

Our approach emphasizes objectivity by focusing on non-depression-specific behavioral markers—head motion dynamics—that are observable even during a patient’s first clinical interaction. Unlike methods requiring longitudinal tracking or disorder-specific symptom coding, we analyze motion patterns inherently linked to anxiety, ensuring applicability in real-world scenarios where clinicians may lack prior patient history. By avoiding depression-specific features, our method enhances generalizability, providing a practical tool for rapid anxiety assessment across diverse populations. It could also be applicable to other patient groups and non-clinical populations, further extending its utility. This aligns with clinical workflows, where initial interviews often serve as the primary basis for early diagnosis and treatment planning.

\begin{figure*}[htbp]
  \centering
  \includegraphics[width=0.90\textwidth]{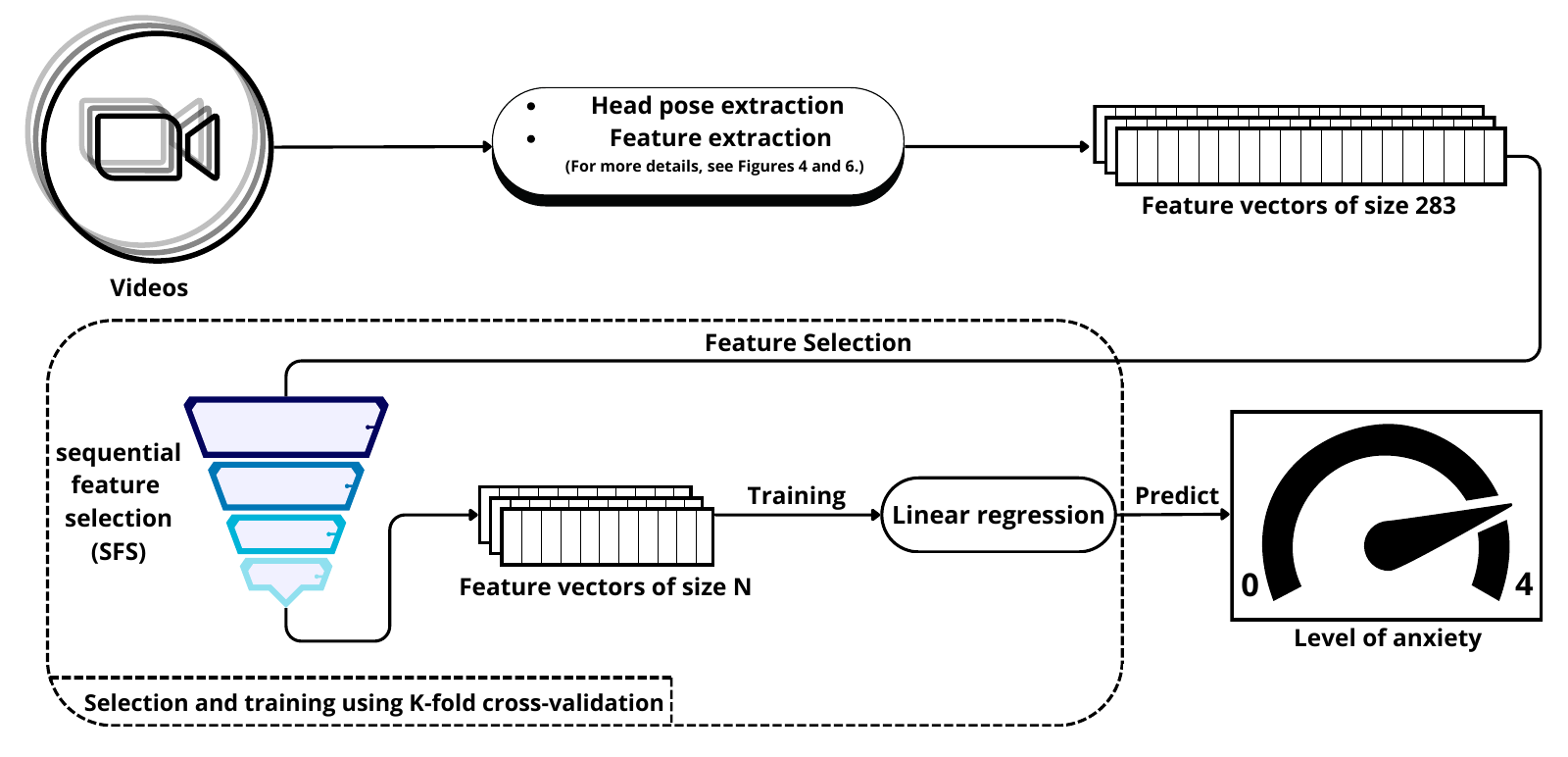}
  \caption{Overview of our proposed pipeline. We apply the same process to all videos of the informal interviews of the CALYPSO dataset. We first extract the head pose and its motion (speed and acceleration) automatically. We then apply statistical feature analysis to extract a feature vector of size 283 for each video. Finally, we apply a cross-validated approach to select features and train a linear model to accurately regress psychological anxiety levels.}
  \label{fig:overview}
\end{figure*}

\section{Related Work}

% Nonverbal cues have been extensively studied for their potential to detect mental health conditions like depression and anxiety. Facial expressions, body gestures, speech patterns, and head movements provide objective insights into a patient's emotional state.

% \subsection{Depression Detection Using Nonverbal Cues}
\subsection{Automatic Depression Detection Using Nonverbal Cues}

Clinicians have observed for a long time that severe depression is correlated with reduced physiological activities, such as monotonic tone, reduced intensity of facial expressions, or low quantity of body motion~\cite{american2013diagnostic}.  

Several studies have indeed shown that modern computer science tools can be used to extract those non-verbal cues and provide an objective way to assess the level of depression. Features extracted from body gestures~\cite{joshi2013can}, speech patterns~\cite{cummins2015review}, facial expressions~\cite{dibekliouglu2017dynamic}, and head movement~\cite{KacemIEEEFG2018} are used to automatically and accurately predict depression severity.
%used to automatically predict the rate of depression severity accurately.
However, the interpretability of most approaches remains unclear. Many recent approaches rely on deep neural networks~\cite{song2018human, suhara2017deepmood}, with improved accuracy but providing limited insights for clinicians.

% Studies like Cohn et al.~\cite{cohn2009detecting} and Girard et al.~\cite{girard2014nonverbal} have shown that certain facial muscle movements and reduced expressiveness correlate with depressive symptoms. 
%Most studies on detecting depression focus on separating severely depressed and healthy populations. The lack of interpretability makes it unclear whether the proposed features could be directly used to assess the precise state of depressed patients.
Most studies on detecting depression focus on separating severely depressed and healthy populations but lack interpretability. Gahalawat et al.~\cite{GahalawatDep} addressed this issue by proposing an interpretable approach using head motion patterns for binary classification. However, while their method enhances explainability, it remains focused on distinguishing between groups rather than evaluating the precise state of depressed patients.

\subsection{Automatic Anxiety Detection}

Similar to depression, research has shown that several anxiety disorders can be automatically detected using non-verbal cues~\cite{lima2019facial}.
Most studies have primarily focused on physiological signals—such as heart rate variability, skin conductance, and cortisol levels—using wearable devices~\cite{physiological2024,hickey2021smart}.

In~\cite{pediaditis2015extraction}, the authors extract facial features, such as face and mouth motion, to predict whether patients are anxious or relaxed. This analysis was later extended in~\cite{giannakakis2017stress}. Mo et al.~\cite{mo2023sff} propose using facial cues to accurately detect anxiety and distress in a non-intrusive manner. However, their feature extraction process relies on deep learning and is therefore non-interpretable.

In our approach, anxiety is considered a symptom of depression, whereas most of the cited works treat anxiety as an independent illness. It remains unclear whether these methods would be effective in detecting anxiety in depressed patients.

\subsection{Tracking behavior with head motion}

Head movements are significant nonverbal indicators for behavioral analysis and are well studied, as they are easy to track in virtual reality environments~\cite{lindner2021better} or dyadic interactions~\cite{won2014automatic}. In virtual reality settings, head movements serve as valuable features for analyzing social interactions~\cite{herrera2021virtual}, emotional states~\cite{xue2021investigating}, and simulation sickness in virtual environments~\cite{bailenson2006longitudinal}. 

By extracting head motion dynamics from videos of structured Hamilton interviews, Kacem et al.~\cite{KacemIEEEFG2018} classified depression severity, finding that depressed individuals exhibit less head movement. Dibeklioglu et al. ~\cite{DibekliogluICMI2015} combined head movements with facial dynamics and vocal prosody for depression detection, noting differences in nodding frequency and amplitude between depressed and non-depressed populations. Finally, head movements have also been shown to be valuable for anxiety prediction~\cite{won2016identifying}. 
\\
\par We summarize the main contributions of our work below:
\begin{enumerate}
\item We introduce the CALYPSO dataset, a longitudinal study of clinical depression. In particular, we propose to use the videos of informal interviews of the study to analyze anxiety in severe depression using head movements extracted from the videos. %CALYPSO is a database of people suffering from severe depression, and in particular, we captured their head movements.
\item We propose segmenting videos into head-moving and non-moving phases. This approach allows us to extract a more comprehensive set of head motion features for our analysis. We further select the most significant features and train a regression model to predict psychological anxiety.
\item We validate our method on the CALYPSO dataset and demonstrate its effectiveness in daily life settings by applying it to informal interviews with depressed patients. For the first time, we establish a link between anxiety in severe depression and objectively measurable nonverbal cues, based on head motion characteristics extracted from pose angular displacements.

\end{enumerate}
%\subsection{Gap in the Literature}

\section{Methodology}
%%%%% je suis arrivé ici %%%%
In this section, we outline our methodology. Our approach is based on videos of informal interviews from the CALYPSO dataset. We divide our approach into several steps. 
First, we automatically extracted the head pose and detailed its motion from the videos. Then, we extracted statistical features from the poses and motion sequences. Our final step is a trainable pipeline to select features and train a linear, interpretable model to predict physical anxiety from the selected features.  

\begin{figure}[htbp]
  \centering
  \includegraphics[height=0.20\textheight]{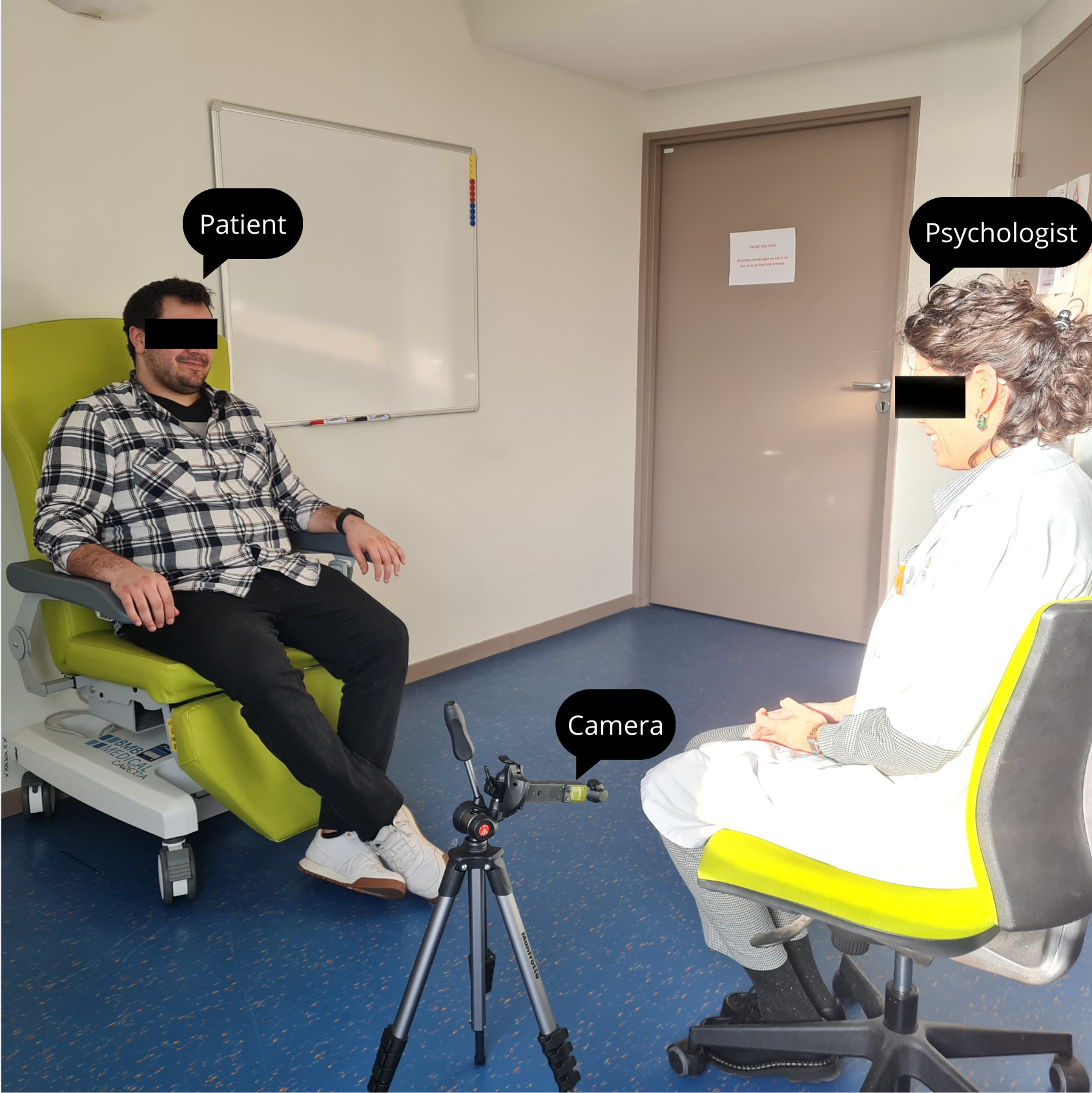}
  \caption{Interview Room Setup for the Calypso Depression Dataset.}
  \label{fig:setup_room}
\end{figure}

\subsection{Head pose and motion extraction}

\noindent \textbf{Pose Extraction}

\noindent To track head orientation in 3D space throughout the video, we used MediaPipe software~\cite{Mediapipe}. We captured the head orientation using Euler angles—pitch, yaw, and roll—representing the different axes of rotation.

\begin{figure}[htbp]
  \centering
  \includegraphics[width=0.30\textwidth]{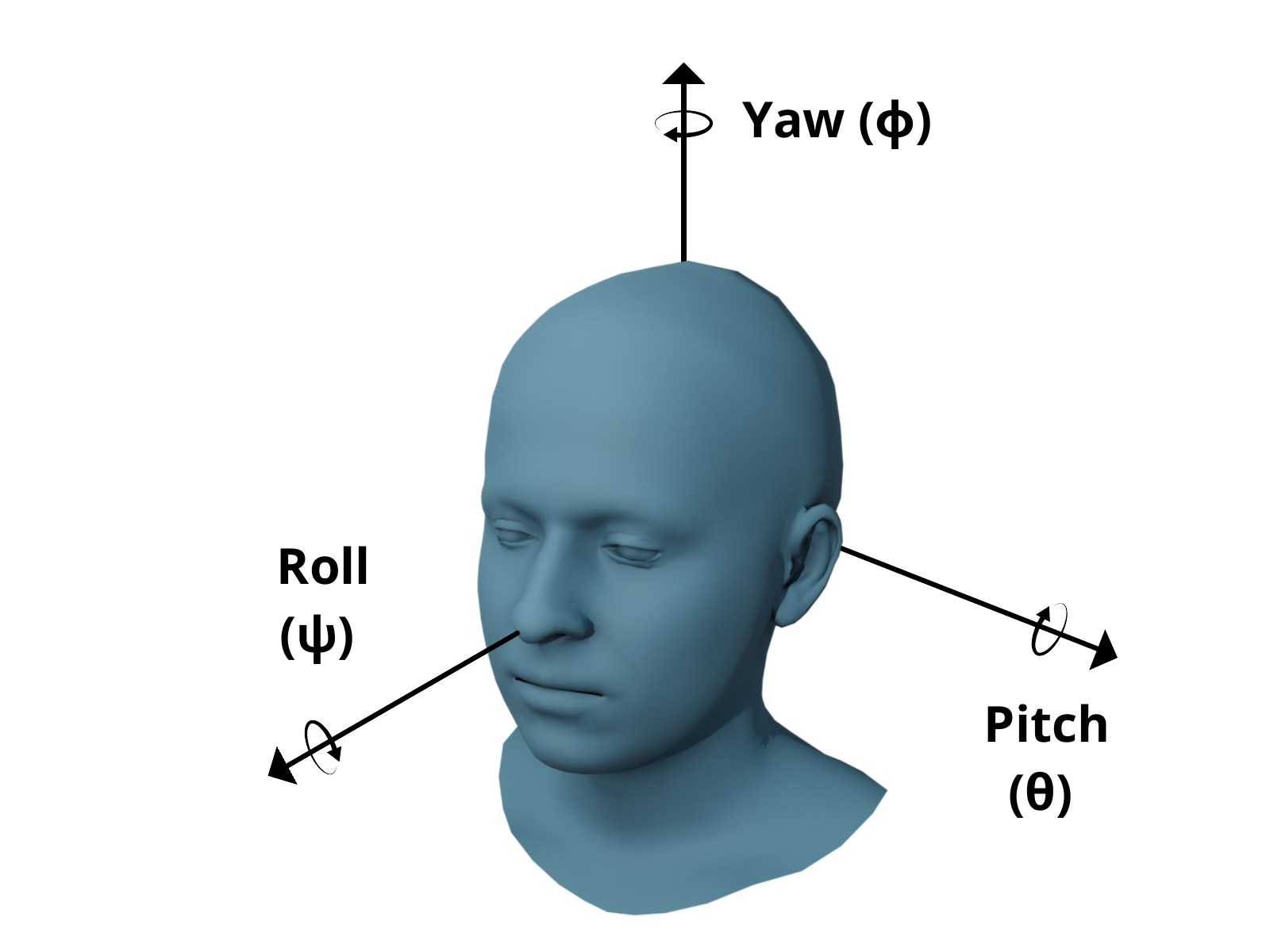}
  \caption{Diagram of head motion axes—pitch, roll, and yaw—used in our analysis.}
  \label{fig:axes_diagram}
\end{figure}

As shown in the figure~\ref{fig:axes_diagram}, the three primary axes of rotation are:
\begin{itemize}
    \item \textbf{Pitch} (\(\theta\)): Rotation around the x-axis (nodding up and down).
    \item \textbf{Yaw} (\(\phi\)): Rotation around the y-axis (turning left and right).
    \item \textbf{Roll} (\(\psi\)): Rotation around the z-axis (tilting the head side to side).
\end{itemize}
This process resulted in a time series of angles that describe the head's orientation throughout the interview, as shown in Figure 4.

\noindent \textbf{Angular Velocity Calculation}

\noindent To measure head motion, we calculated the angular velocity for yaw, pitch, and roll.
Let t be a time step and $t+\Delta t$ be the following step. As the pose can be described as a rotation matrix (computed from the yaw pitch and roll angles), we compute the derivative of the rotation matrix $R_t$. 

This derivative is computed using the following formula:
\[
\omega(t) \approx \frac{1}{\Delta t} \left( R_{t+\Delta t} R_{t}^T - I \right).
\]

The product \( R_{t+\Delta t} R_{t}^T \) is the relative rotation between two consecutive poses, and we measure its deviation from the identity matrix \( I \). The resulting matrix \(\omega(t)\) is a skew-symmetric matrix, from which we can recover the angular velocity across yaw, pitch, and roll axes:

\[
\omega(t) = 
\left[
\begin{array}{ccc}
0 & -\omega_z & \omega_y \\
\omega_z & 0 & -\omega_x \\
-\omega_y & \omega_x & 0
\end{array}
\right]
\]

where \(\omega_x\) is the angular velocity around the pitch-axis, \(\omega_y\) is the angular velocity around the yaw-axis, and \(\omega_z\) is the angular velocity around the roll-axis.

\begin{figure*}[htbp]
  \centering
  \includegraphics[width=0.99\textwidth]{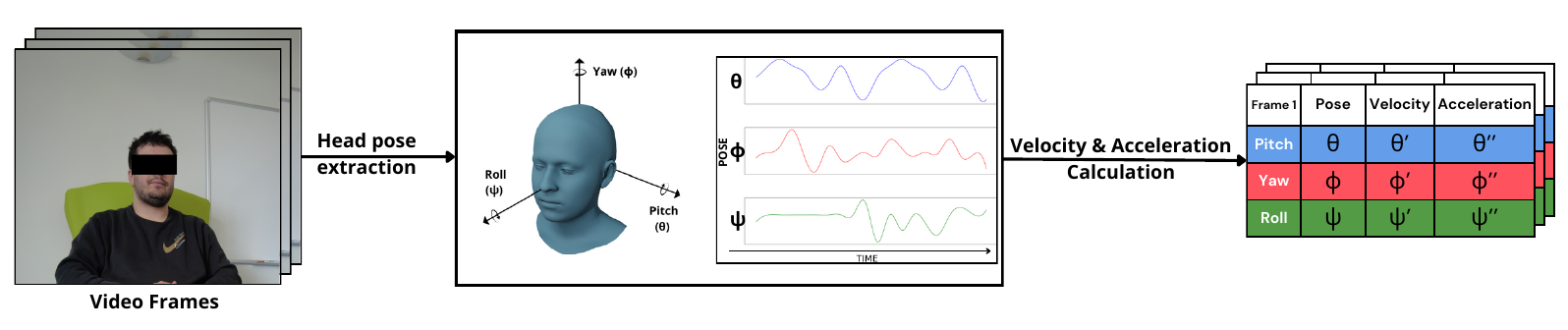}
  \caption{Illustration of the head pose and motion extraction.}
  \label{fig:pose_extraction}
\end{figure*}

\noindent \textbf{Acceleration Calculation}

\noindent Since the skew-symmetric matrix space is linear, calculating the angular acceleration is simply derived from:

\[
\dot{\omega}(t) \approx \frac{\omega_{t+1} - \omega_{t}}{\Delta t}
\]

where \( \omega_{t} \) and \( \omega_{t+1} \) are the angular velocities at consecutive timestamps \( t \) and \( t+1 \), and $\Delta{t}$ is the time interval between the measurements.

The pitch, yaw, and roll acceleration are defined as $\dot{\omega}_x, \dot{\omega}_y, \dot{\omega}_z$.

\subsection{Motion segmentation}

\noindent We observed that statistical features extracted from the full interview sequences are not discriminative enough to provide reliable predictions (see Table 1). Moreover, we noted that the head pose of the patient alternates between two states: \textbf{moving} (the patient is changing position on the chair) and \textbf{steady} (the patient is on a stable position and has limited motion). This motivated us to segment interviews in moving and steady sequences.

To provide a flexible yet straightforward method for achieving this, the velocity data was clustered into two groups using a Gaussian Mixture Model (GMM). The two clusters are effective at classifying head movement between the intuitive "moving" and "steady" states.

\begin{figure}[htbp]
  \centering
  \includegraphics[width=0.47\textwidth]{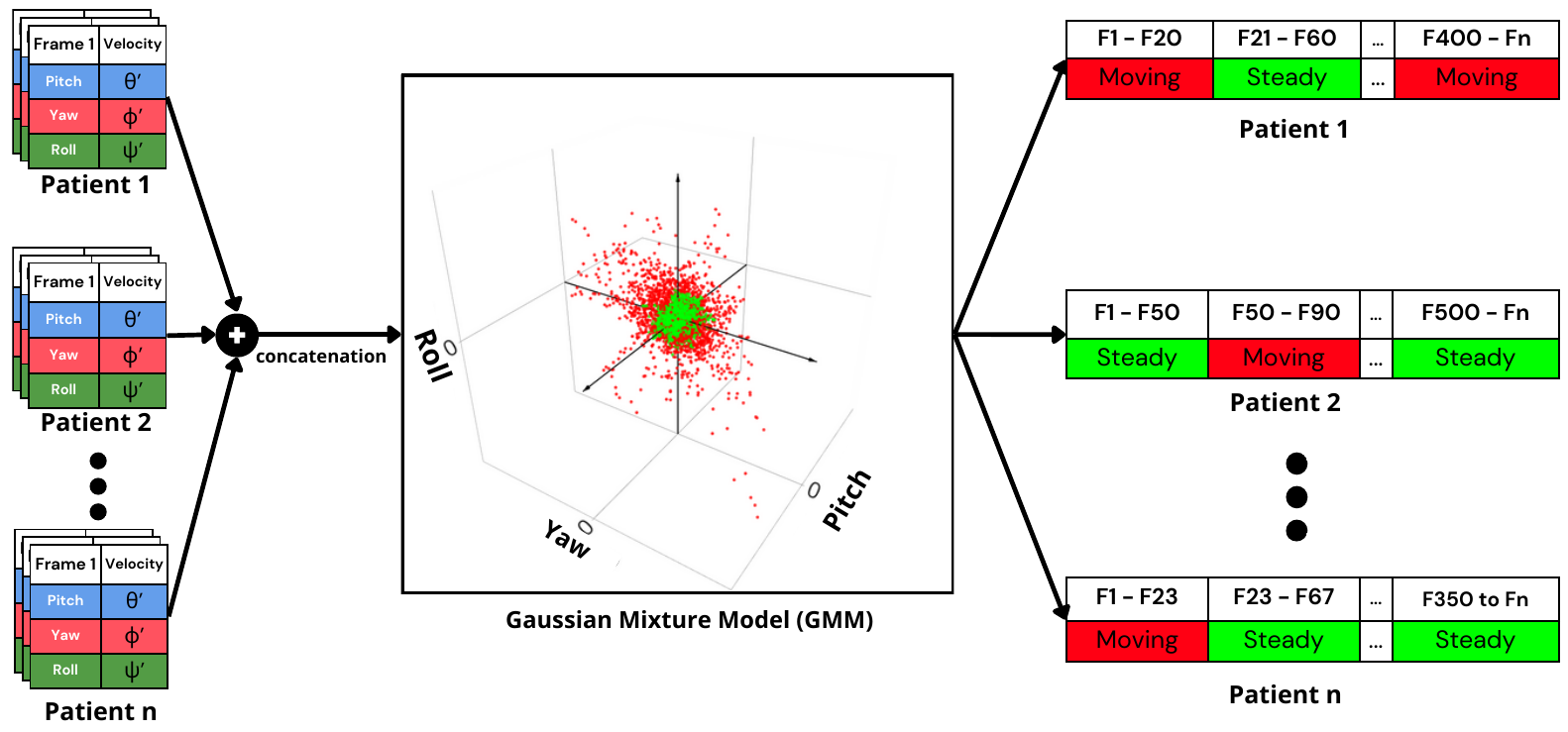}
  \caption{We apply a Gaussian Mixture Model (GMM) to cluster head rotational velocities (pitch, yaw, roll) into "moving" and "steady" states, segmenting interviews into sequences (e.g., F21-F60, representing frames 21 to 60). The plot illustrates the velocity profiles for each axis.}
  \label{fig:gmm_clusters}
\end{figure}

This approach can be generalized across different patients, making the classification robust for various head movement behaviors. The whole process is illustrated in Figure~\ref{fig:gmm_clusters}.

\subsection{Feature Extraction}

\begin{figure*}[htbp]
  \centering
  \includegraphics[width=0.95\textwidth]{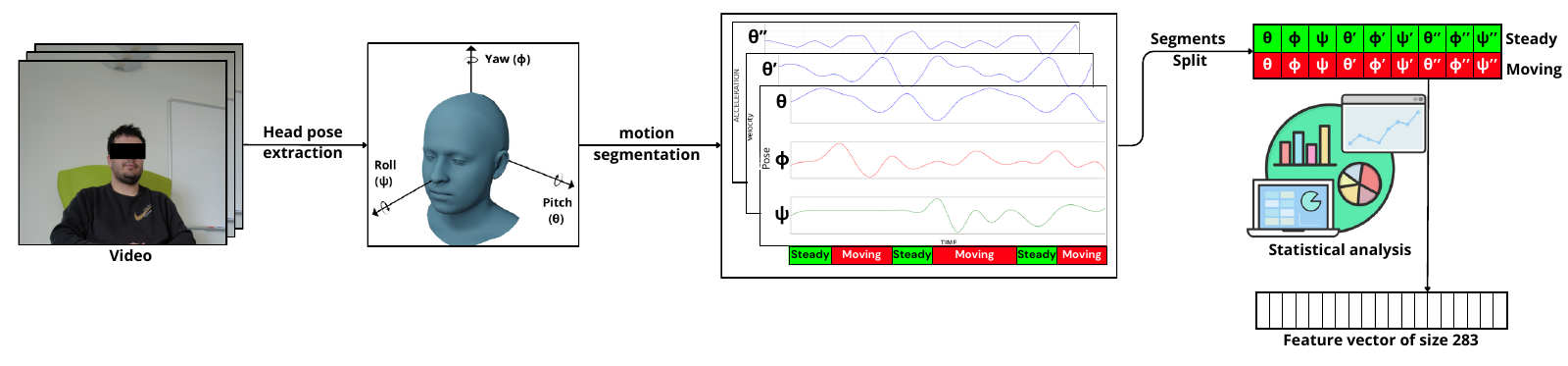}
  \caption{Illustration of the full feature extraction process.}
  \label{fig:feature_extraction}
\end{figure*}

Using the resulting clustering from the Gaussian Mixture Model (GMM), the interview videos were segmented into moving and steady sequences. For each segment, head movements were analyzed in terms of pitch, yaw, and roll, along with their velocities and accelerations. This resulted in a total of nine core features: three rotational angles (pitch, yaw, roll), their velocities, and their accelerations. (9 features across 2 clusters)

To comprehensively capture the characteristics of head movement, we extracted three types of statistical features:
\begin{itemize}
    \item \textbf{Global statistical features:}  
    For a general overview, we grouped the “moving” and “stable” segments into two subsets and calculated summary statistics, including Mean, Median, Range, Median Absolute Deviation (MAD), Skewness, Kurtosis, and Standard Deviation for each feature. This resulted in a total of 7 statistics for 9 features across 2 clusters (moving and stable).  ($7\times 9 \times 2$)
    \item \textbf{Sequence-Level Features:} We then separated each moving and steady sequence and analyzed them individually. We then computed the following  7 statistics: Mean, Median, Range, MAD, Skewness, Kurtosis, and Standard Deviation of pitch, yaw, roll, velocities, and accelerations. These features were then averaged across all sequences in each group (moving or steady) ($7 \times 9 \times 2$). Additionally, we included the sum of absolute values for all speeds and accelerations (cumulative displacement, cumulative acceleration). ($1 \times 6 \times 2$).

    \item \textbf{Temporal Features:}  
    For the temporal analysis, we focused on the duration of each movement or stable segment. We calculated the mean, median, standard deviation, skewness, range of durations, and the ratio of time spent in each state (moving vs. stable). Additionally, we computed the number of transitions per minute, which reflects how often the subject switched between movement and stillness. ($6 \times 2$ + 1)

\end{itemize}

This process resulted in a feature vector of size 283 ( 126 global, 144 sequence-level, and 13 temporal) for each video. 

\subsection{Feature Selection and Model Development}

To reduce the risk of overfitting given the high number of features, we applied a selection method that reduces the number of features. Such approaches have shown useful in extracting interpretable features for depression assessment~\cite{9253541, 10042796}.

\noindent \textit{Correlation Filtering}  
To ensure that the model was not impacted by multicollinearity, and to simplify the process, we first computed the correlation between all extracted features. Any features with a correlation coefficient greater than ±0.8 were removed (with a preference for non-derivative features). This step reduced redundancy and eliminated high correlation in features that could affect the reliability of the model. After applying this filtering process, we retained a refined set of \textbf{96 features}.

We then employed Sequential Feature Selection (SFS), which allowed us to identify the most relevant features for our task. Sequential Feature Selection is a feature selection technique that iteratively adds or removes features based on their contribution to the model's performance. In our approach, we explored two variations of SFS:
\par \textit{Sequential Selection by Exclusion:}
This method sequentially removes the least significant feature based on a predefined performance criterion (e.g., Mean Squared Error (MSE)). At each iteration, the feature whose exclusion results in the least degradation or the most improvement in model performance is removed. This process continues until no features are left. We then select the set of features that provided the best performance during the process.

\par \textit{Sequential Backward Floating Selection (Inclusion/Exclusion):}
This approach adds a conditional inclusion step to backward selection. After removing a feature, the algorithm evaluates whether reintroducing any of the previously excluded features can enhance the model's performance. This mechanism thus dynamically adjusts the feature subset, which allows recovering from suboptimal exclusions and identifying a more optimal set of features. 

\begin{figure}[htbp]
  \centering
  \includegraphics[width=0.47\textwidth]{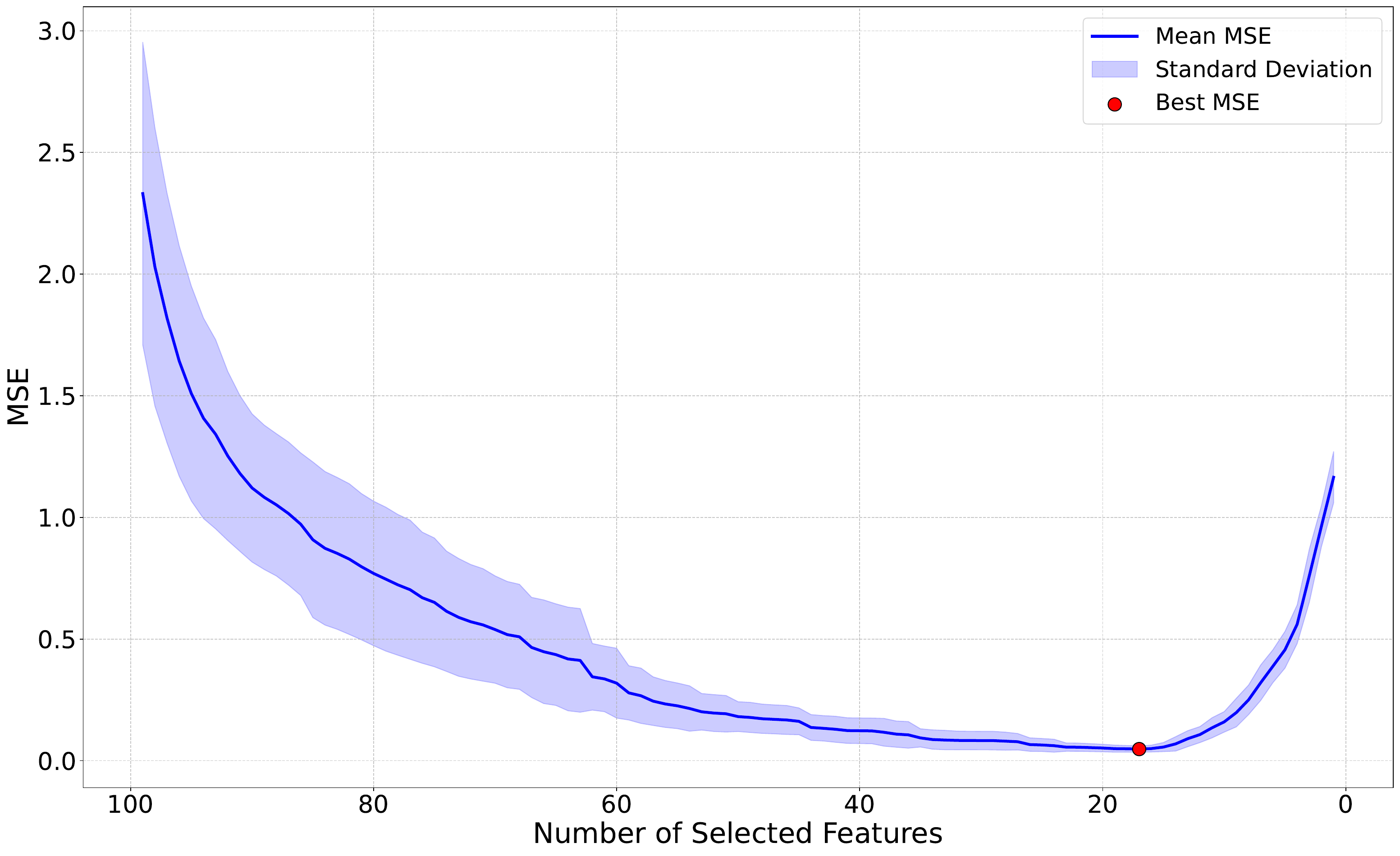}
  \caption{The plot shows the cross-validation (CV) scores during feature selection across 10 folds. The blue line represents the mean CV score as the number of selected features increases, with the shaded area indicating the standard deviation. The x-axis shows the number of selected features, and the y-axis shows the mean squared error (MSE). The red marker indicates the mean of the lowest CV scores across all folds.}
  \label{fig:features_selection}
\end{figure}

\subsection{Regression model}
We used a linear regression model to allow for interpretation of the model behavior. We used the Lasso regularization as it consistently yielded the best results. All steps using training data as input were included in the cross-validation process to ensure our results were not overfitted or exhibiting spurious correlations. 

\section{RESULTS}

In this section, we present the outcomes of our regression analysis aimed at predicting anxiety levels based on head movement features. We evaluate multiple machine learning models under different feature selection processes and assess the impact of incorporating motion segmentation. We finally provide an interpretation of our model behavior and features that are shown to be linked with the presence or absence of anxiety in severe depression.

\subsection{CALYPSO dataset}

\par \noindent \textit{Patient Selection}
\par \noindent Patients admitted to the hospital undergo standard diagnostic evaluations with an attending psychiatrist. A clinician then conducts a first examination to determine if the patient's history and symptoms align with the DSM-5 (Diagnostic and Statistical Manual of Mental Disorders) conditions for severe clinical depression~\cite{american2013diagnostic}. Patients who meet the inclusion criteria are proposed for a more in-depth interview, contributing directly to the CALYPSO depression dataset. In total, \textbf{32 patients} meeting these criteria were included in the study. The CALYPSO clinical trial has been reviewed and approved by the Ethics Committee under approval number 2022-A01160-43, ensuring adherence to ethical standards and guidelines. All patients provided written informed consent before participation in the study.

The study participants were predominantly French nationals (ethnicity data were not collected), with an equal gender distribution (50\% male, 50\% female).

\par \noindent \textit{Interview Process}
\par \noindent Selected patients participated in structured clinical interviews conducted in a controlled environment, as shown in Figure~\ref{fig:setup_room}. The interview was divided into two distinct phases: an initial informal segment that consists of a casual conversation between the psychologist and the patient, lasting only a few minutes, during which the patient is asked non-medical questions. 

\begin{figure}[htbp]
  \centering
  \includegraphics[width=0.47\textwidth]{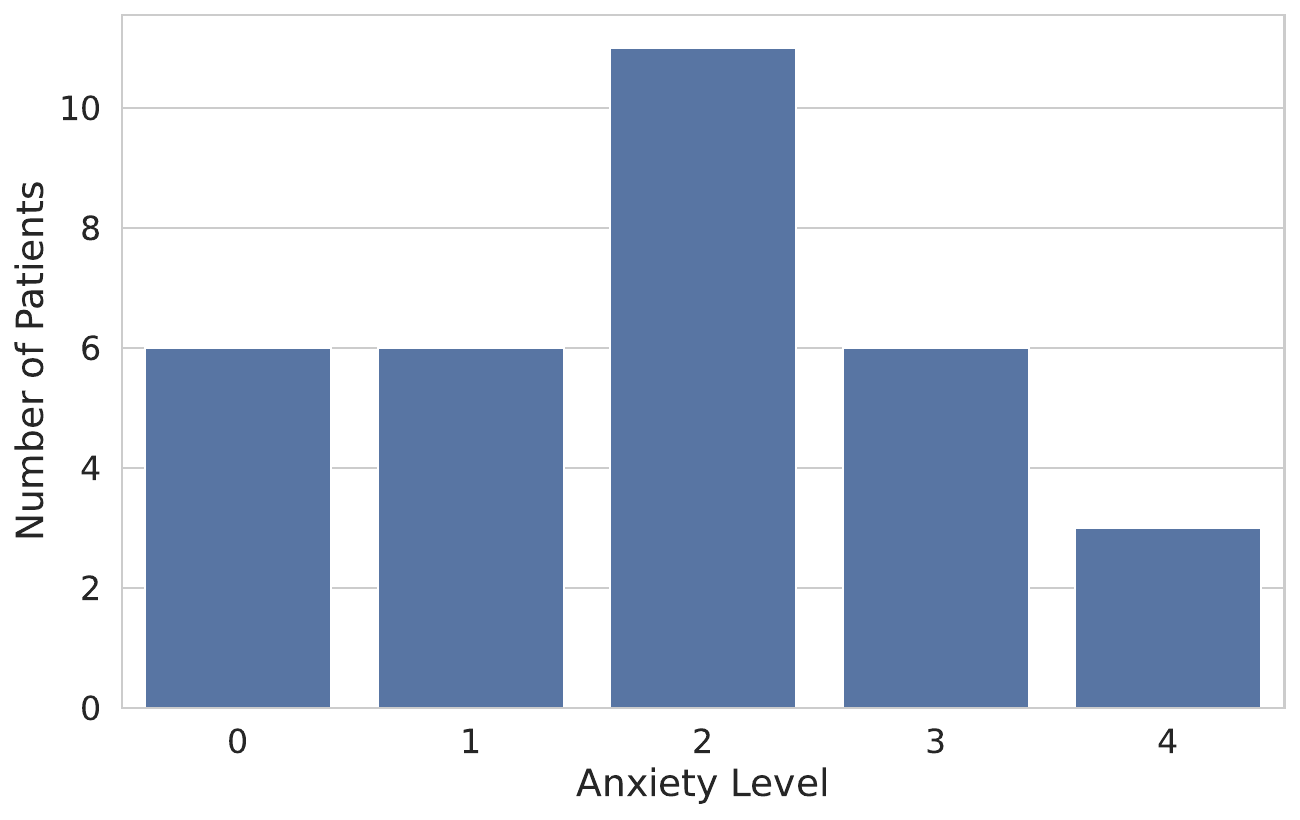}
  \caption{Distribution of Anxiety Levels in the CALYPSO Dataset.}
  \label{fig:anxiety_distribution}
\end{figure}

After the informal conversation, the clinician conducts a structured interview aimed at evaluating the Hamilton Depression Rating Scale (HDRS) for the patient. This assessment specifically includes measuring psychological anxiety. After the interview, the clinician assigns a psychological anxiety score ranging from 0 (no anxiety) to 4 (high anxiety), based on the patient's responses and observed behaviors, as shown in Figure~\ref{fig:anxiety_distribution}.

%Following this, the structured interview is designed to provide the Hamilton Depression Rating Scale. In particular, the score of psychological anxiety, rating between 0 and 4 is calculated at the end of the interview. 

To assess if the head motion patterns can be used in daily life interviews, such as discussions with a general practitioner, we conducted our analysis on the video recordings from the informal part of the interview.

\subsection{Experimental setup}

To identify the most effective set of features and minimize the risk of overfitting, we employed Sequential Feature Selection (SFS) in combination with 10-fold cross-validation. We evaluated multiple machine learning models and fine-tuned the alpha parameter to determine the optimal configuration for our dataset. In each cross-validation fold, SFS was applied to the training data (comprising 9 folds) to select the best-performing features, resulting in 10 distinct feature lists. We then consolidated these results by selecting features that appeared in at least five of the ten lists, ensuring that only the most consistently important features were retained. This approach enhanced the model's robustness and generalizability by focusing on reliable predictors. Finally, we trained the final model using this refined set of features, which streamlined the model and improved its performance on unseen data.

\subsection{Evaluation Metrics}

The regression performance was evaluated using the mean absolute error (MAE);

\begin{equation}
\text{MAE} = \frac{1}{n} \sum_{i=1}^{n} \left| y_i - \hat{y}_i \right|
\end{equation}

and the coefficient of determination (R2 score),:

\begin{equation}
R2 = 1 - \frac{\sum_{i=1}^{n} (y_i - \hat{y}_i)^2}{\sum_{i=1}^{n} (y_i - \bar{y})^2}
\end{equation}
where, $n$ is the number of observations, $y_i$ is the actual value, $\hat{y}_i$ is the predicted value, and $\bar{y}$ is the mean of the actual values.

To further assess the practical applicability of our regression model, we converted the regressed predictions into discrete values. Each predicted value is converted to the closest integer. We then report the classification accuracy based on the predicted anxiety level.

\subsection{Psychological Anxiety Level Prediction}
\begin{figure}[htbp]
  \centering
  \includegraphics[width=0.40\textwidth]{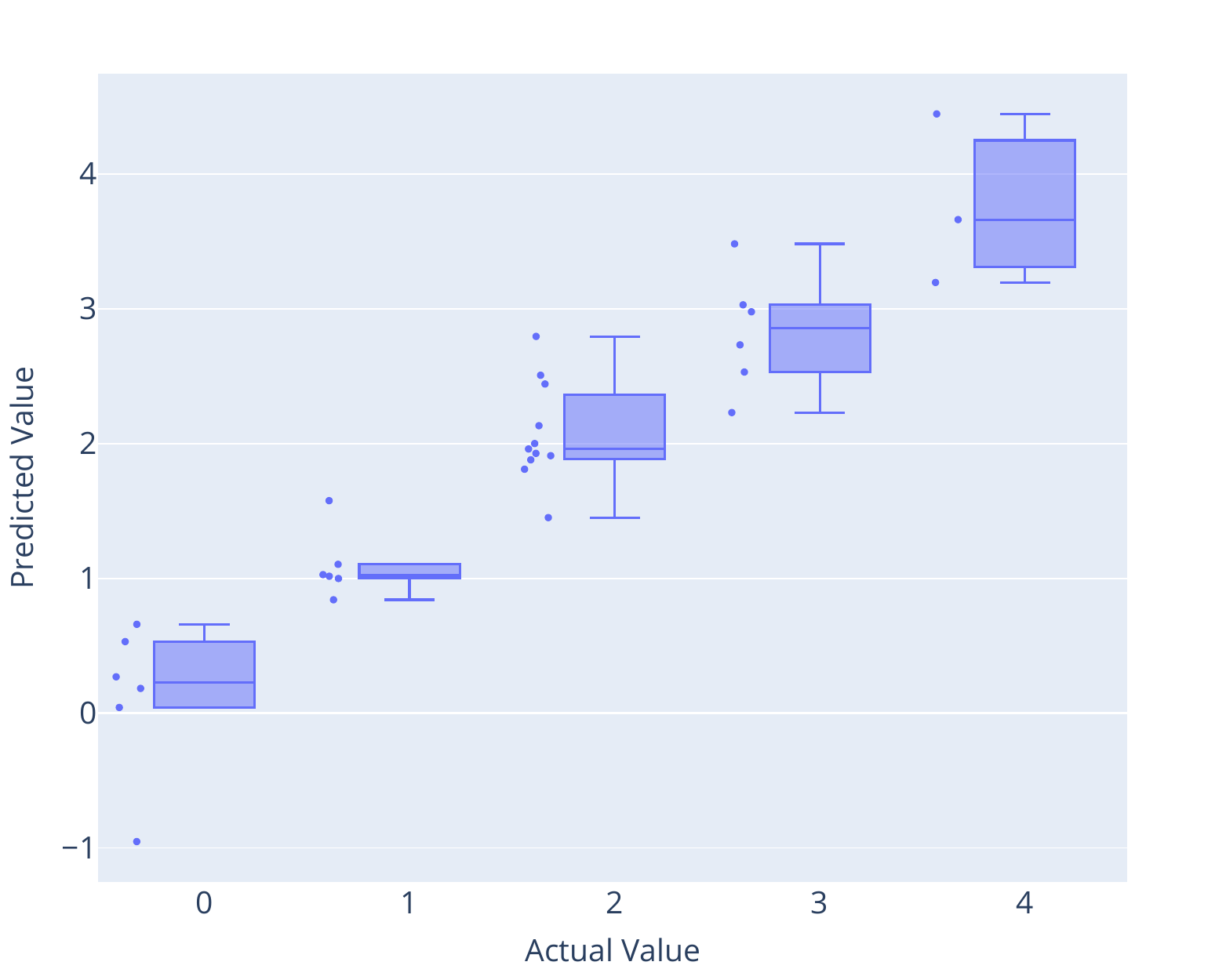}
    \caption{This figure illustrates the actual versus predicted psychological anxiety levels for the best Lasso model, each point represents a patient}
    \label{fig:results_psychological}
\end{figure}

For predicting psychological anxiety levels, the best-performing model was Lasso regression, achieving a Mean Absolute Error (MAE) of 0.31 and a coefficient of determination ($R2$) of 0.87 with only 14 features needed (see Figure \ref{fig:results_psychological}). 

\subsection{Impact of segmentation}
To evaluate the significance of the interview motion segmentation in our feature extraction process, we compared it to a baseline model without the segmentation. The baseline approach involved extracting features directly from the raw head motion data without segmentation. We computed the same statistical measures (mean, median, range, skewness, kurtosis, standard deviation) for the head angles (pitch, yaw, roll), velocities, and accelerations, for a total of 54 features.

As shown in Table~\ref{tab:with_without_gmm}, our approach significantly improves the predictive performance of the model compared to the baseline.

\begin{table}[htbp]
\caption{Regression Results for Psychological Anxiety Level}
\label{tab:with_without_gmm}
\centering
\begin{tabular}{|l|c|c|c|c|}
\hline
\textbf{Selection Process} & \textbf{Features} & \textbf{MAE} & \textbf{R2}  & \textbf{Accuracy} \\
\hline
\textbf{Full model (ours)} & 14 & \textbf{0.31} & \textbf{0.87} & \textbf{0.75} \\
\hline
\textbf{Without GMM} & 16 & 0.90 & 0.10 & 0.44 \\
\hline
\end{tabular}
\end{table}

\subsection{Comparison of Feature Selection Methods}
We also compared different feature selection methods to evaluate their impact on model performance. Specifically, we examined exclusion-only selection, inclusion and exclusion (I/E) selection, and no selection at all. Table~\ref{tab:selection_comparison} summarizes the results.

\begin{table}[htbp]
\caption{Comparison of Feature Selection Methods for Psychological Anxiety Level}
\label{tab:selection_comparison}
\centering
\resizebox{0.47\textwidth}{!}{%
\begin{tabular}{|l|c|c|c|c|}
\hline
\textbf{Selection Process} & \textbf{No. of} & \textbf{MAE} & \textbf{R2} & \textbf{Accuracy}\\
 & \textbf{Features} &  &  &  \\
\hline

\textbf{Exclusion Only} & 12 & 0.48 & 0.76 & 0.50 \\
\hline
\textbf{Inclusion and Exclusion (I/E)} & \textbf{14} & \textbf{0.31} & \textbf{0.87} & \textbf{0.75} \\
\hline
\textbf{I/E - Without GMM} & 16 & 0.90 & 0.10 & 0.44 \\
\hline
\textbf{No selection} & 283 & 0.98 & -0.05 & 0.31 \\
\hline
\end{tabular}
} % End of resizebox
\end{table}

The results indicate that the inclusion and exclusion (I/E) feature selection method yields the best performance. Notably, the exclusion only process results in fewer features than the inclusion and exclusion strategy. Moreover, the Lasso $L_1$ penalty alone is not sufficient to select the features without overfitting or satisfying performance (R2 score is almost zero, meaning that the model always predicts the mean value).

\subsection{Classification Performance Analysis}

To provide a qualitative analysis of the classification, we present the confusion matrix of our model in Figure ~\ref{fig:Confusion_matrix}.

\begin{figure}[htbp]
  \centering
  \includegraphics[width=0.40\textwidth]{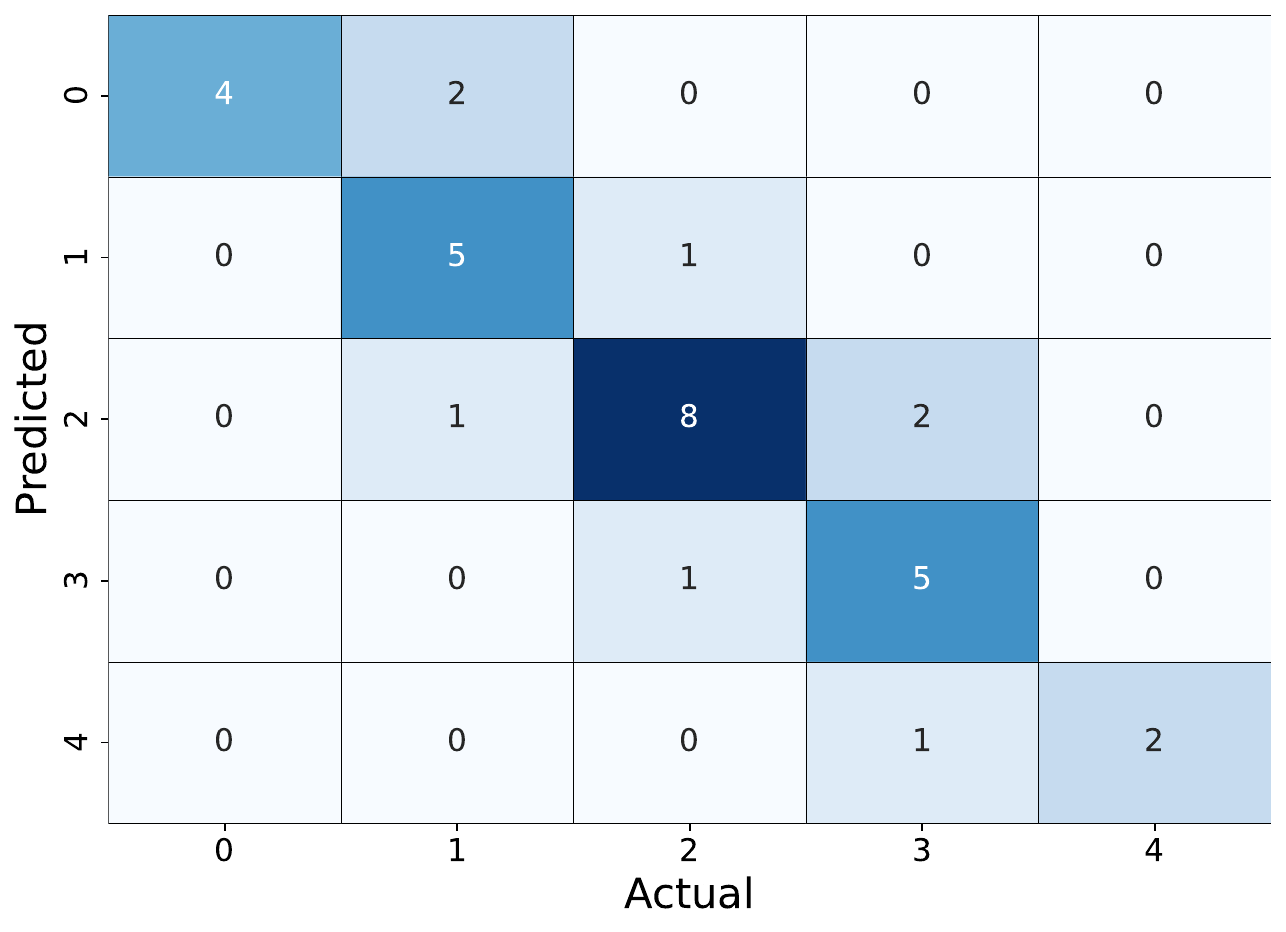}
    \caption{Confusion Matrix: Classification results by grouping continuous anxiety level predictions into classes with a tolerance of ±0.5 units.}
    \label{fig:Confusion_matrix}
\end{figure}

We observe that all wrong predictions differ by one, which we find acceptable given the inherent variability in psychiatrist ratings. This result suggests that our model's performance is comparable to human-level error margins in clinical settings.

\subsection{Interpretation of results}

\begin{figure}[htbp]
  \centering
  \includegraphics[width=0.45\textwidth]{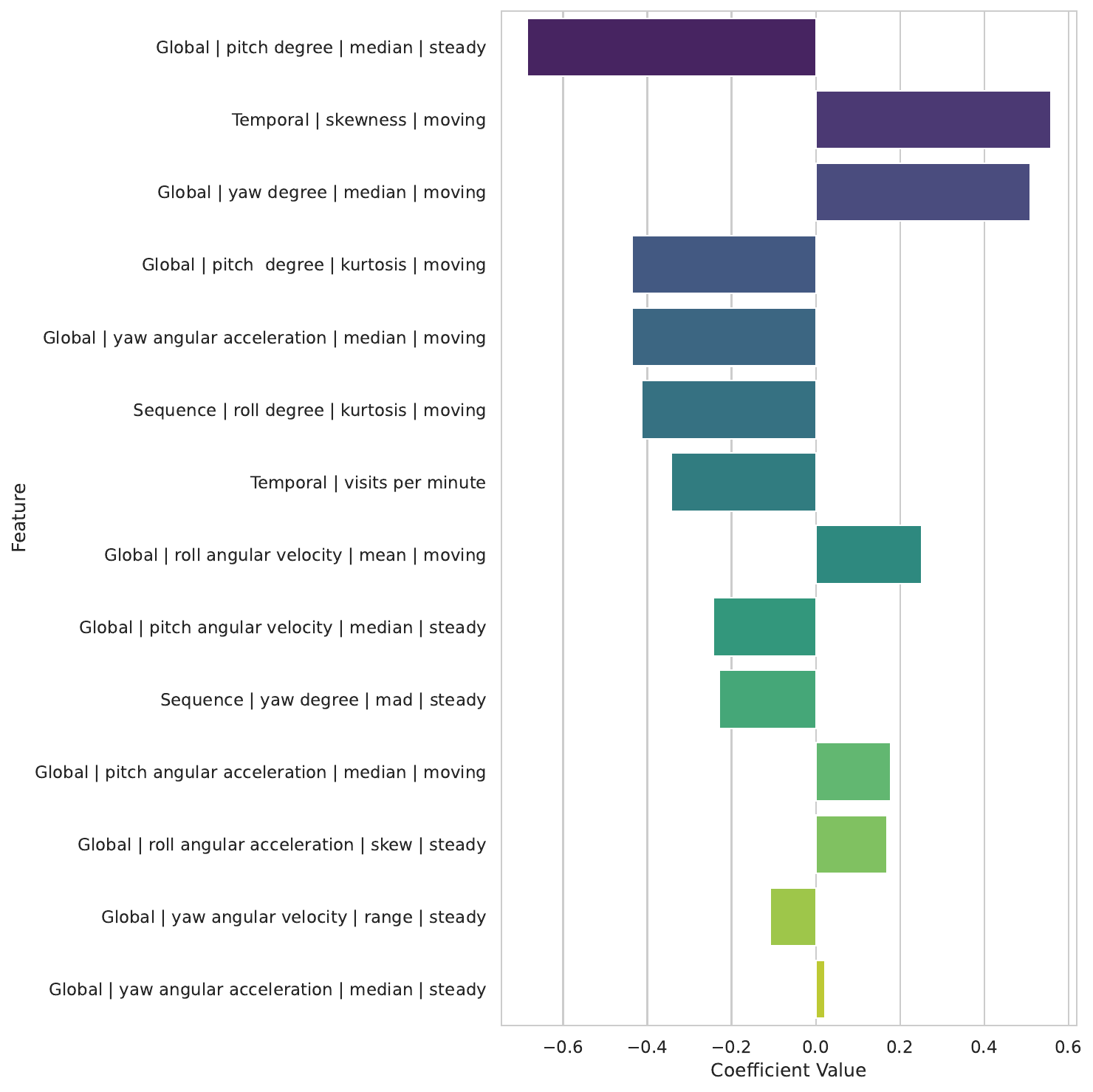}
    \caption{This plot illustrates the coefficients of the Lasso regression model for each feature in the best model.}
    \label{fig:lasso_coef_GMM}
\end{figure}

Figure \ref{fig:lasso_coef_GMM} presents the best model coefficients derived from the regression analysis predicting anxiety levels based on head movement data. Each bar represents the importance or weight of a specific feature, with the feature names listed on the y-axis and their respective coefficient values on the x-axis. The key observations include:

\begin{itemize}
    \item \textbf{Global | Pitch Degree | Median | Steady}: This feature shows the highest negative coefficient, indicating that steady pitch movements (up and down head movements) are inversely linked to anxiety levels. In other words, when the median pitch degree is low (indicating the head is raised), anxiety levels tend to be higher.
    
    \item \textbf{Temporal | Skewness | Moving}: The second-highest coefficient (positive) indicates that patients with high anxiety tend to spend longer periods in sequences where they are moving. Skewness highlights the imbalance in time spent during movement. In other words, patients with a high level of physical anxiety have an irregular duration of motion, and alternate between long moments of motion, and shorter ones, compared to more stability for non anxious patients.

    \item \textbf{Temporal | Visits per Minute}: This feature measures the number of transitions between moving and steady periods per minute during the observation period. It correlates negatively with anxiety levels, which might seem counterintuitive at first. However, this suggests that more anxious patients tend to either stay moving for long periods or remain still for extended durations, resulting in fewer transitions between motion and steady states. By combining information from other features, we deduce that higher anxiety levels are associated with patients staying in motion for longer stretches, while less anxious patients frequently alternate between moving and stopping.
\end{itemize}

\begin{figure}[htbp]
  \centering
  \includegraphics[width=0.45\textwidth]{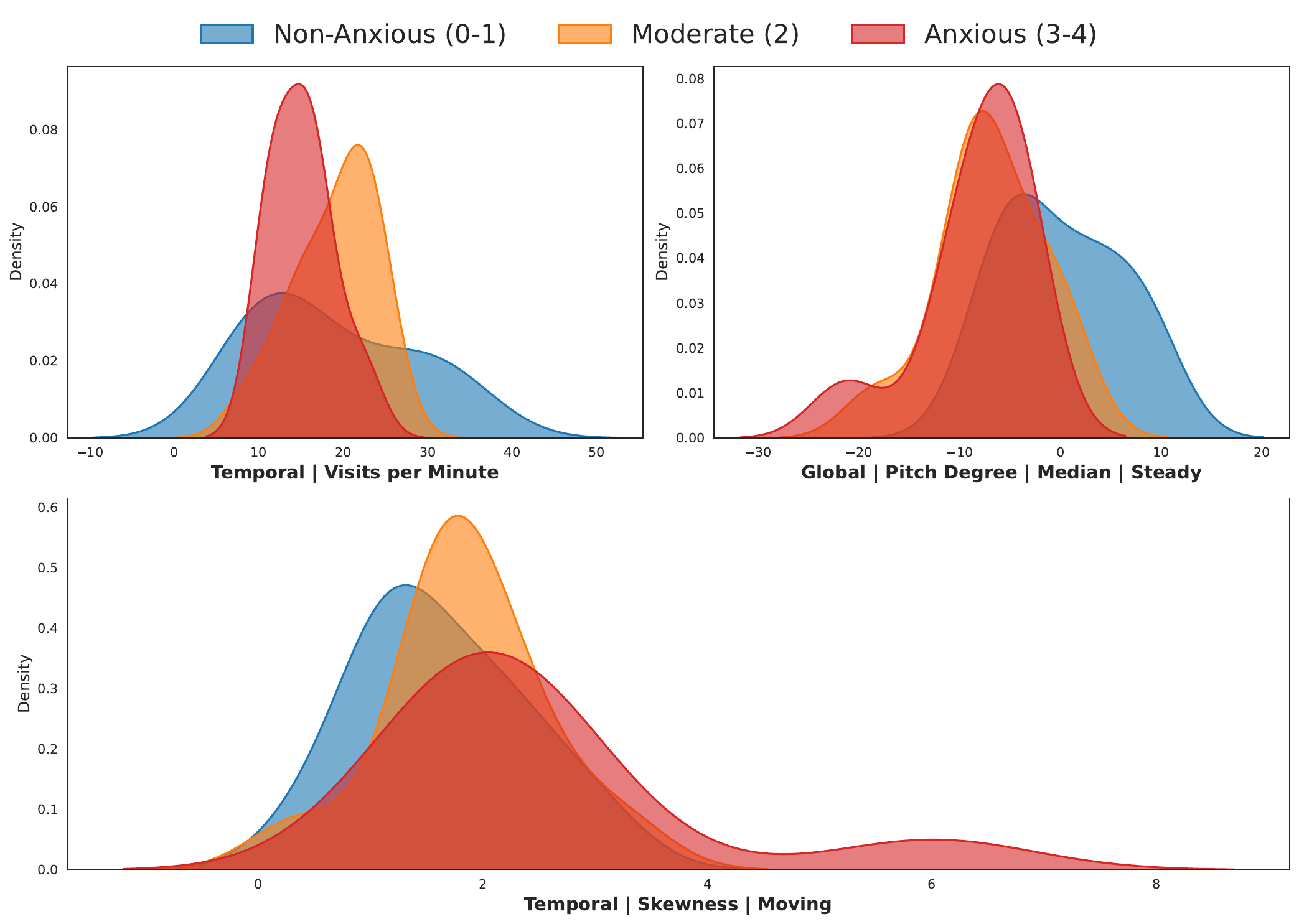}
    \caption{Density distributions of key features across different anxiety score groups. The blue, orange, and red areas represent the density estimates for individuals with anxiety scores of 0-1 (non-anxious), 2 (moderately anxious), and 3-4 (highly anxious), respectively.}
    \label{fig:densities}
\end{figure}

The density plots in Figure \ref{fig:densities} illustrate how these key features vary across different anxiety levels, offering a more detailed insight into the trends that correspond with the model's predictions. 

Overall, the model suggests that stable head motion is linked to lower anxiety, while unstable head motion, particularly with longer and more irregular motion sequences, is associated with higher anxiety.

%While a single feature might not always provide clear insight, understanding multiple features in combination gives a comprehensive overview, allowing us to better interpret their contribution to the overall prediction.

\section{Limitations and future work}

\subsection{Limitations}
%Fouad tu peux discuter des résultats sur l'anxiété somatique?
While our method demonstrates strong performance in predicting psychological anxiety levels using head motion patterns, it is less effective in assessing somatic anxiety. We applied the same technique to provide a regression model for somatic anxiety levels. However, the regression models yielded lower predictive accuracy, with a Mean Absolute Error (MAE) of \textbf{0.47} and an R² score of \textbf{0.53}. With these results, the large error obtained does not allow for a clear differentiation between the affected and non-affected populations.

Somatic anxiety manifests through physical symptoms such as muscle tension, restlessness, and other bodily sensations that may not be fully captured by analyzing head movements alone. We believe those manifestations often involve subtle physiological changes or whole-body movements that require additional modalities to detect accurately. Incorporating other physiological or behavioral cues may be necessary to comprehensively assess this subtype of anxiety in individuals with severe depression.

\subsection{Future Works}

Several avenues of research are possible to validate and expand upon this work.

    %\item \textbf{Clustering Head Positions:} Since our results indicate that specific head positions are significant in assessing anxiety levels, future work could involve clustering head positions to identify common postural patterns associated with anxiety. This approach may reveal distinct head orientation profiles that correlate with varying degrees of anxiety, enhancing the interpretability of our model.

    %\item \textbf{Combining with Facial Landmarks:} Utilizing facial landmark data can capture subtle facial expressions and micro-expressions associated with anxiety. By combining head motion analysis with facial feature tracking, we can develop a more comprehensive assessment tool that considers a wider range of non-verbal cues.
    
    %\item \textbf{Incorporating Eye Gaze Analysis:} Integrating eye gaze tracking can provide additional insights into anxiety-related behaviors. Eye movements, such as fixation patterns and saccades, are known to be influenced by emotional states and could enhance the predictive power of our model when combined with head movement data.

    %\item \textbf{Increasing Dataset Size:} Expanding the CALYPSO Depression Dataset with more participants will enhance the generalizability of our results. Collaborating with additional clinical centers can facilitate data collection from a more diverse patient population, reducing potential biases arising from a limited sample size.

\par \textbf{Multimodal Analysis:} This study focuses on head motion patterns. However, CALYPSO dataset contains full videos of patients. Integrating other features, from facial expressions, body gestures or speech patterns could help provide measurements of depression symptoms. Combining multimodal data may improve the predictive accuracy of the model and offer deeper insights into the behavioral manifestations of anxiety in depression.

\par \textbf{Longitudinal Studies:} In this study, our goal was to develop a model that accurately predicts anxiety levels in depressed patients based on their first clinical interview. However, future work will explore applying this approach to longitudinal data to monitor anxiety evolution throughout treatment. Investigating whether changes in head motion patterns correlate with treatment response could provide valuable insights for adjusting therapeutic strategies over time.

\section{Conclusion}
In this study, we introduced a novel, non-invasive method for quantifying anxiety severity in patients with severe depression by analyzing head motion patterns during clinical interviews. We propose also a new depression dataset named CALYPSO, introduced in this paper, which contains video data of depressed patients, from which we extracted features related to head motion. Moreover, our new approach, separating moving and non-moving segments of the interview, allows us to extract more valuable features for the analysis. We demonstrated that we can train an interpretable model based on the selected features for predicting the anxiety level of depressed patients, achieving a Mean Absolute Error (MAE) of 0.31 and an $R^2$ of 0.87.

These results suggest that head motion patterns can serve as reliable, objective indicators of anxiety severity in individuals with severe depression. By providing an automated and quantifiable assessment tool, our approach has the potential to assist psychiatrists in making more informed decisions regarding diagnosis and treatment planning. This method enhances the understanding of anxiety's role in depression and contributes to more personalized and effective interventions for patients suffering from both conditions.

%%%%%%%%%%%%%%%%%%%%%%%%%%%%%%%%%%%%%%%%%%%%%%%%%%%%%%%%%%%%%%%%%%%%%%%%%%%%%%%%

%{\small
%\bibliographystyle{ieee}
%\bibliography{egbib}
%}

%%%
\section{Appendix}

\subsection{Confusion matrix of somatic Anxiety}

We show in Figure~\ref{fig:Confusion_matrix} the confusion matrix of our approach regression model for somatic anxiety. Notably, the moderately anxious class is hardly well predicted compared to the results of psychological anxiety in the main paper, supporting the need for more non-verbal features to assess the somatic anxiety symptom.

\begin{figure}[htpb]
  \centering
  \includegraphics[width=0.9\linewidth]{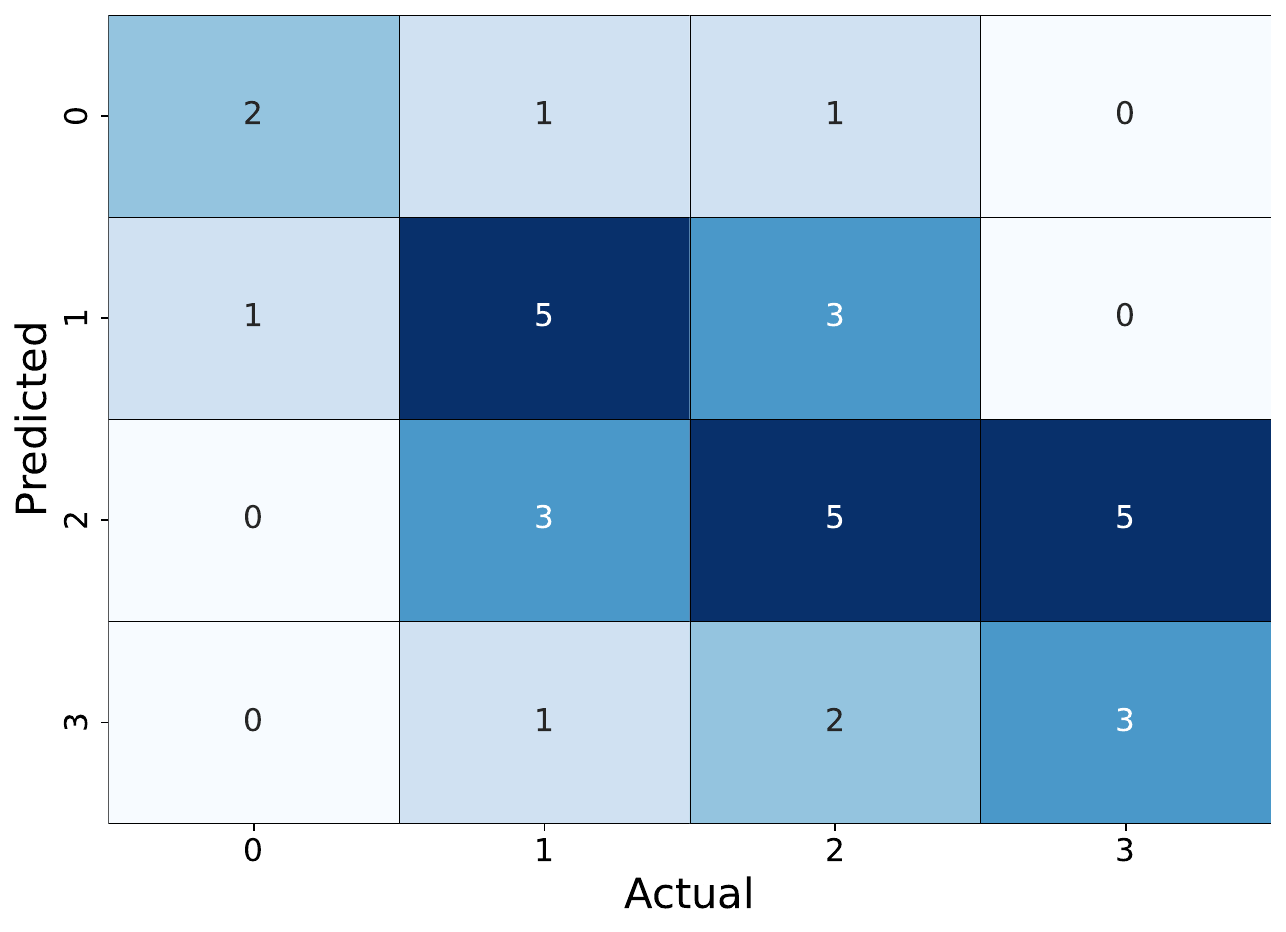}
  \caption{Confusion Matrix obtained using the best model from Table~\ref{tab:fused_results_somatic}. Classification results are shown by grouping continuous anxiety level predictions into classes with a tolerance of $\pm$0.5 units.}
  \label{fig:Confusion_matrix}
\end{figure}

\subsection{Supplementary results for prediction of psychological anxiety}

We provide in Table \ref{tab:fused_results}, a detailed ablation of each pipeline parameter. In particular, we show that the Lasso model is the best regularization for regressing psychological anxiety.

\begin{table*}[htbp]
\centering
\caption{Full ablation study of our model for psychological anxiety. The best model is the Lasso model with I/E selection process.}
\label{tab:fused_results}
\begin{tabular}{|p{3.5cm}|l|c|c|c|c|}
\hline
\textbf{Selection Process for SFS} & \textbf{Model} & \textbf{Alpha}$^{\mathrm{a}}$ & \textbf{Number of Features Selected}$^{\mathrm{b}}$ & \textbf{MAE}$^{\mathrm{c}}$ & \textbf{R2}$^{\mathrm{c}}$ \\
\hline
\textbf{Exclusion Only} & \textbf{Ridge} & 1.0   & 23 & \textbf{0.35} & \textbf{0.87} \\
                        &                 & 0.1   & 16 & 0.49          & 0.76 \\
\cline{2-6}
                        & \textbf{Lasso}  & 0.01  & 12 & 0.48          & 0.76 \\
\cline{2-6}
                        & \textbf{Linear Regression} & N/A   & 23 & 0.53          & 0.64 \\
\cline{2-6}
                        & \textbf{ElasticNet} & 0.01  & 13 & 0.59          & 0.65 \\
\hline \hline
\textbf{Inclusion and Exclusion} & \textbf{Ridge} & 1.0   & 18 & 0.46          & 0.78 \\
                                 &                & 0.1   & 16 & 0.49          & 0.76 \\
\cline{2-6}
                                 & \textbf{Lasso} & 0.1   & 6  & 0.63          & 0.61 \\
                                 &                & 0.01  & 14 & \fbox{\textbf{0.31}} & \fbox{\textbf{0.87}}\\
                                 &                & 0.001 & 16 & 0.63          & 0.60 \\
\cline{2-6}
                                 & \textbf{Linear Regression} & N/A   & 13 & 0.72          & 0.45 \\
\cline{2-6}
                                 & \textbf{ElasticNet} & 0.01  & 9  & 0.57          & 0.68 \\
                                 &                      & 0.001 & 14 & 0.60          & 0.67 \\
\hline \hline
\textbf{Inclusion and Exclusion (Without GMM)} & \textbf{Ridge} & 1.0   & 17 & 0.80         & 0.35 \\
                                               &                & 0.1   & 24 & \textbf{0.67}        & \textbf{0.45} \\
\cline{2-6}
                                               & \textbf{Lasso} & 0.01  & 16 & 0.90      & 0.10 \\
\cline{2-6}
                                               & \textbf{Linear Regression} & N/A   & 30 & 1.6     & -2.45 \\
\cline{2-6}
                                               & \textbf{ElasticNet} & 0.01  & 7 & 0.76 & 0.42 \\
\hline
\end{tabular}

\end{table*}

% \small
% \item \textbf{Note:} Values in \fbox{boxed cells} indicate the best results across all selection processes. \textbf{Bold} values indicate the best results within each selection process.
% \item $^{\mathrm{a}}$ Regularization parameter.
% \item $^{\mathrm{b}}$ Number of features selected by the model.
% \item $^{\mathrm{c}}$ MAE: Mean Absolute Error; R2: Coefficient of Determination.

\subsection{Supplementary results for prediction of somatic anxiety}

We provide in Table \ref{tab:fused_results_somatic}, a similar ablation for somatic anxiety. Notably, the Lasso model fails to provide similar accuracy. Moreover, no model is able to provide similar results as for psychological anxiety.

\begin{table*}
\centering
\caption{Full ablation study of our model for psychological anxiety. The best model is the ElasticNet model with Exclusion selection process}
\label{tab:fused_results_somatic}
\begin{tabular}{|p{3.5cm}|l|c|c|c|c|}
\hline
\textbf{Selection Process for SFS} & \textbf{Model} & \textbf{Alpha}$^{\mathrm{a}}$ & \textbf{Number of Features Selected}$^{\mathrm{b}}$ & \textbf{MAE}$^{\mathrm{c}}$ & \textbf{R2}$^{\mathrm{c}}$ \\
\hline
\textbf{Exclusion Only} & \textbf{Ridge} & 1.0   & 16 & 0.76 & 0.38 \\
                        &                 & 0.1   & 22 & 0.46        & 0.65 \\
\cline{2-6}
                        & \textbf{Lasso}  & 0.01  & 9 & 0.61          & 0.21 \\
\cline{2-6}
                        & \textbf{Linear Regression} & N/A   & 24 & 1.16         & -1.66 \\
\cline{2-6}
                        & \textbf{ElasticNet} & 0.01  & 10 & \fbox{\textbf{0.40}}          & \fbox{\textbf{0.69}} \\
\hline \hline
\textbf{Inclusion and Exclusion} & \textbf{Ridge} & 1.0 & 15 & \textbf{0.47} & \textbf{0.63} \\
                                 &                & 0.1 & 13 & 0.50 & 0.59 \\
\cline{2-6}
                                 & \textbf{Lasso} & 0.01 & 9 & 0.59  & 0.26  \\
                                 &                & 0.001 & 14 & 0.47 & 0.53 \\ 
\cline{2-6}
                                 & \textbf{Linear Regression} & N/A  & 12 & 0.49  & 0.56  \\
\cline{2-6}
                                 & \textbf{ElasticNet} & 0.01 & 7 & 0.52 & 0.47 \\
\hline \hline
\textbf{Inclusion and Exclusion (Without GMM)} & \textbf{Ridge} & 1.0   & 9 & 0.67  & 0.19 \\
                                               &                & 0.1   & 20 & 0.79   & -0.06 \\
\cline{2-6}
                                               & \textbf{Lasso} & 0.01  & 7 & \textbf{0.64}    & \textbf{0.29} \\
\cline{2-6}
                                               & \textbf{Linear Regression} & N/A   & 32 & 2.67          & -12.07 \\
\cline{2-6}
                                               & \textbf{ElasticNet} & 0.01  & 11 & 0.69 & 0.17 \\
\hline
\end{tabular}
\end{table*}
% \small
% \item \textbf{Note:} Values in \fbox{boxed cells} indicate the best results across all selection processes. \textbf{Bold} values indicate the best results within each selection process.
% \item $^{\mathrm{a}}$ Regularization parameter.
% \item $^{\mathrm{b}}$ Number of features selected by the model.
% % \item $^{\mathrm{c}}$ MAE: Mean Absolute Error; R2: Coefficient of Determination.

%%%%%%%%%%%%%%%%%%%%%%%%%%%%%%%%%%%%%%%%%%%%%%%%%%%%%%%%%%%%%%%%%%%%%%%%%%%%%%%%

%%%%

\end{document}